\documentclass[letterpaper, 10 pt, journal, twoside]{IEEEtran}
\ifCLASSINFOpdf
\else
\fi

\usepackage{fontawesome}
\usepackage{cuted}

\usepackage{multirow}
\usepackage{multicol}
\usepackage{booktabs}
\usepackage{graphicx}
\usepackage{float} 
\usepackage{wrapfig}
\usepackage{subcaption}
\usepackage[table]{xcolor}
\usepackage{makecell}

\usepackage{amsmath}
\usepackage{amssymb}
\usepackage{mathtools}

\usepackage{hyperref}
\usepackage[capitalize,noabbrev]{cleveref}

\crefname{algorithm}{Alg.}{Algs.}
\Crefname{algocf}{Algorithm}{Algorithms}
\crefname{section}{Sec.}{Secs.}
\Crefname{section}{Section}{Sections}
\crefname{table}{Tab.}{Tabs.}
\Crefname{table}{Table}{Tables}
\crefname{figure}{Fig.}{Fig.}
\Crefname{figure}{Figure}{Figure}
\Crefname{equation}{Eq.}{Eqs.}
\Crefname{equation}{Equation}{Equations}
\crefname{appendix}{Appx.}{Appx.}
\Crefname{appendix}{Appendix}{Appendix}

\newcommand{\revise}[1]{{#1}}
\newcommand{\final}[1]{{#1}}

\begin{document}
%
\title{Neural Internal Model Control: Learning a Robust Control Policy via Predictive Error Feedback}
%
%
%

\author{Feng Gao$^{1}$, Chao Yu$^{13\dagger}$, Yu Wang$^1$, Yi Wu$^{12\dagger}$
\thanks{This work was supported by National Natural Science Foundation of China (No.62406159, 62325405), Postdoctoral Fellowship Program of CPSF under Grant Number (GZC20240830, 2024M761676), China Postdoctoral Science Special Foundation 2024T170496.} 
\thanks{$^{\dagger}$ Corresponding authors. \texttt{\{zoeyuchao,jxwuyi\}@gmail.com}.}
\thanks{$^{1}$ Tsinghua University, Beijing, China.}%
\thanks{$^{2}$ Shanghai Qi Zhi Institute, Shanghai, China.}
\thanks{$^{3}$ Beijing Zhongguancun Academy, Beijing, China.}
}

%
%

\markboth{IEEE Robotics and Automation Letters. Preprint Version.} 
{Gao \MakeLowercase{\textit{et al.}}: Neural Internal Model Control}

%



\maketitle

\begin{abstract}
Accurate motion control in the face of disturbances within complex environments remains a major challenge in robotics. Classical model-based approaches often struggle with nonlinearities and unstructured disturbances, while reinforcement learning (RL)-based methods can be fragile when encountering unseen scenarios. In this paper, we propose a novel framework, Neural Internal Model Control (NeuralIMC), which integrates model-based control with RL-based control to enhance robustness. Our framework streamlines the predictive model by applying Newton-Euler equations for rigid-body dynamics, eliminating the need to capture complex high-dimensional nonlinearities. This internal model combines model-free RL algorithms with predictive error feedback. Such a design enables a closed-loop control structure to enhance the robustness and generalizability of the control system. We demonstrate the effectiveness of our framework on both quadrotors and quadrupedal robots, achieving superior performance compared to state-of-the-art methods. Furthermore, real-world deployment on a quadrotor with rope-suspended payloads highlights the framework's robustness in sim-to-real transfer. Our code is released at \url{https://github.com/thu-uav/NeuralIMC}.
\end{abstract}


\section{Introduction}
\label{sec:intro}

\IEEEPARstart{A}{chieving} stable and precise motion in complex, disturbed environments is essential for robots, such as quadrotors maintaining stable flight in wind fields and legged robots carrying loads over diverse terrains. This challenge, known as the robust control or adaptive control problem, has been extensively studied~\cite{acbook}. Classical adaptive control methods adjust control parameters in real-time to compensate for dynamic environmental changes, often using mathematical models of the system and disturbances~\cite{nguyen2018model,rivera1986internal,aastrom1973self,lin2000self}, while robust control methods utilize more informative signals to enhance robustness, like Internal Model Control~\cite{francis1976internal}. Despite their guaranteed stability, these methods rely on accurately modeling system dynamics, which is challenging for nonlinear systems with unstructured disturbances.
In contrast, reinforcement learning (RL) methods have shown remarkable success in continuous control across various robot morphologies, including skill learning for armed robots~\cite{2018-TOG-deepMimic,li2023learning}, high-speed quadrotor racing~\cite{kaufmann2023champion,song2023reaching}, and adaptive locomotion for legged robots~\cite{doi:10.1126/scirobotics.aau5872,kumar2021rma,long2023him,li2024reinforcement,radosavovic2024real}. Large-scale training with randomized dynamics and environmental parameters enables RL-based control methods to generalize across different conditions and environments~\cite{peng2017simtoreal,rudin2021learning}, allowing robots to perform reliably even in previously unseen and highly varying scenarios.

\begin{figure}[t]
    \centering
    \includegraphics[width=0.9\linewidth]{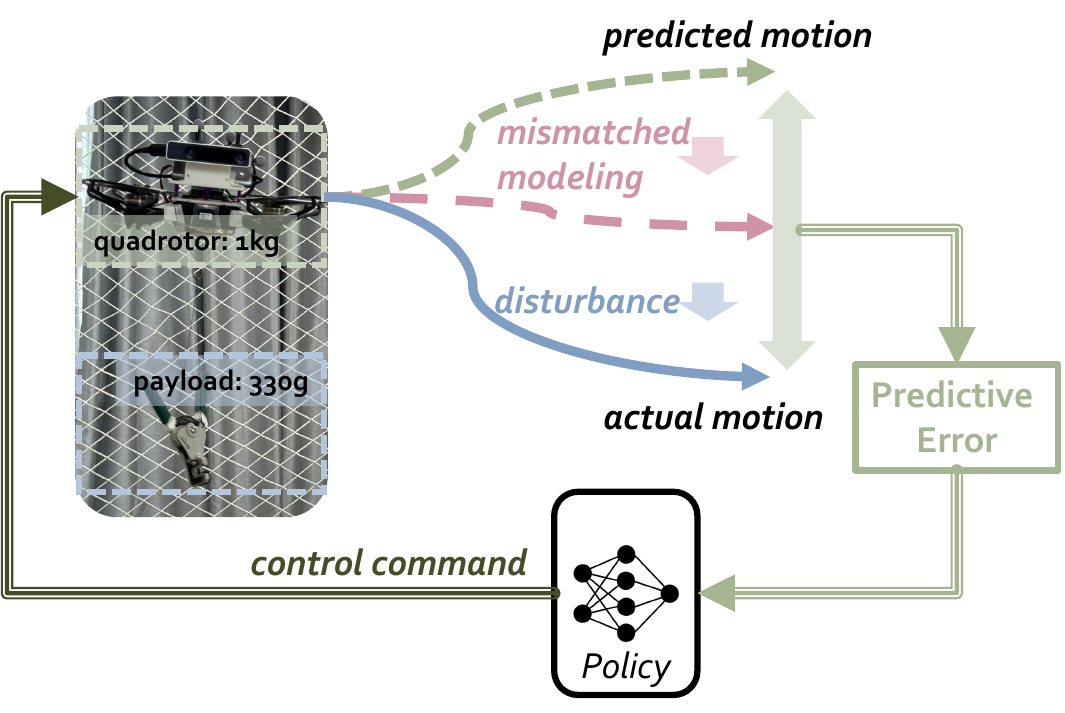}
    \caption{Predictive errors can be caused by mismatched system modeling and/or external disturbances.}
    \label{fig:pef}
    \vspace{-2mm}
\end{figure}

There has been a recent trend towards integrating model-based control with RL-based neural control to harness the strengths of both worlds~\cite{o2022neural,li2022bridging,long2023him,huang2023datt}. 
One representative framework is to use predicted future states or estimated disturbances as an additional input signal to obtain a more informative model-free RL policy. 
Long et al.~\cite{long2023him} employ unsupervised contrastive representation learning to train a predictive encoder capable of estimating the next state in the latent space. Conversely, Huang et al.~\cite{huang2023datt} use a handcrafted $\mathcal{L}_1$ adaptive law~\cite{l1ac} to explicitly estimate applied force disturbances, resulting in a robust trajectory tracking controller that can handle unexpected wind fields. 
Although promising, integrating model-based and RL-based controls faces significant challenges. Predictive models often struggle with variability when training data cannot cover real-world dynamics, and handcrafted disturbance estimators require precise tuning, limiting their feasibility for complex systems and adaptability to dynamic changes.
For establishing a more general method, we need to develop a structure that effectively combines these controls to fully leverage their advantages and address these limitations.

\begin{figure*}[t]
    \centering
    
    \vspace{1.2mm}
    \includegraphics[width=0.75\textwidth]{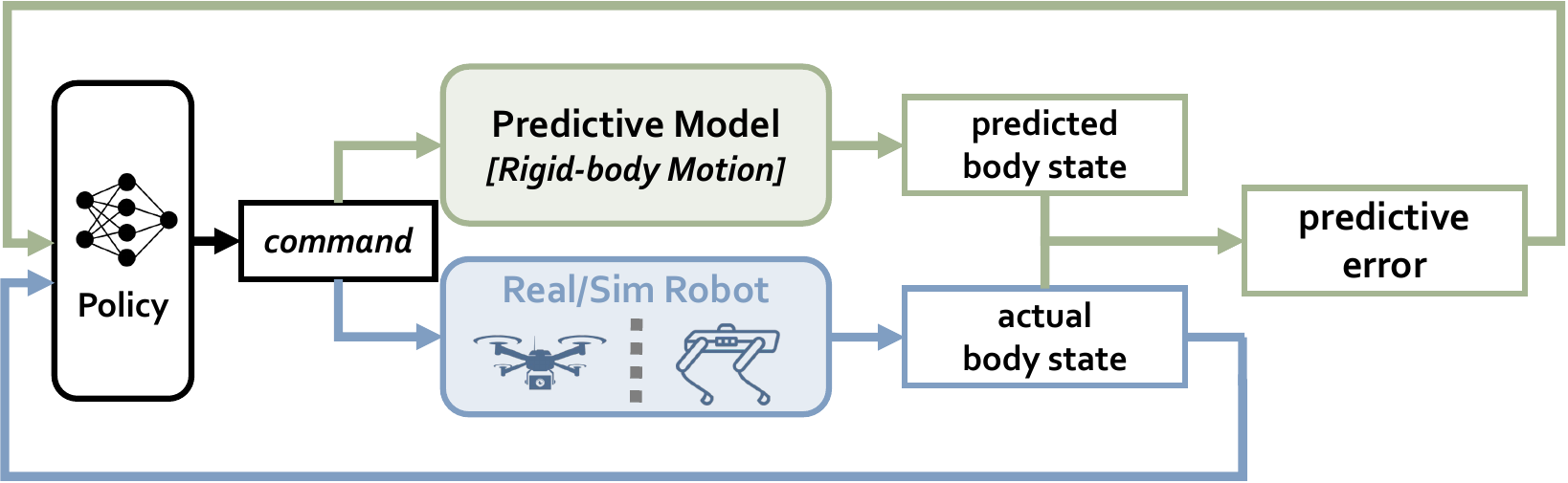}
    \caption{The closed-loop NeuralIMC framework using an explicit predictive model to calculate the next body state, applirope to both quadrotors  and legged robots.}
    \vspace{-1mm}
    \label{fig:overview}
\end{figure*}

\revise{In this paper, we propose a new learning-based control methods which leverages a predictive error feedback to enhance control robustness, inspired by model-based control}.
By simplifying the model to focus on the robot's main body and treating its dynamics as a rigid transformation, we bypass the high-dimensional nonlinearity of complex robot systems and environmental interactions. 
This simplification enables us to use an explicit predictive model without requiring parameter tuning, establishing a foundation for robust closed-loop control. We compute the predictive error—defined as the discrepancy between plant and model responses—as a feedback signal to the policy, enhancing robustness and generalizability through informative feedback and extensive training. Our framework, shown in \cref{fig:overview}, can be easily integrated with other learning algorithms by incorporating this feedback loop.
To demonstrate the effectiveness and versatility of our method, we apply it to two prevalent robot morphologies: quadrotors, which demand precise 3D control, and quadrupedal robots, which require adaptive locomotion to navigate complex terrain interactions. Empirical results indicate that our framework can surpass recent state-of-the-art control methods in both domains. Furthermore, we conduct a real-world deployment on a quadrotor with various rope-suspended payloads to illustrate its robustness in sim-to-real transfer.

\section{Related Work}
\label{sec:related}

\subsection{Classical control and disturbance estimation}

Classical adaptive controllers require estimating system parameters in a closed-loop manner to adjust control actions in real-time. $\mathcal{L}_1$ adaptive control has been widely used to estimate force disturbances in quadrotors~\cite{mallikarjunan2012l1,pravitra2020L1}. Neural-Fly~\cite{o2022neural} incorporated pre-trained representations to enhance rapid online adaptation through deep learning, achieving stable flight in previously unseen wind conditions. 
Internal Model Control (IMC)~\cite{francis1976internal,rivera1986internal} is a general structure for robust control that uses a predefined model to predict system responses and estimate disturbances. However, applying IMC to more complex and nonlinear systems is challenging due to the difficulty of crafting reliable models for such systems. \revise{Hybrid Internal Model (HIM) further leverages a learning-based internal model that employs unsupervised contrastive learning, with the learned model acting as a predictive encoder for estimating system responses in the latent space, as proposed by Long et al.~\cite{long2023him}.} Despite their advantages, these methods either rely on explicit disturbance estimation or are heavily influenced by prediction accuracy, which can significantly affect performance if compromised.

\subsection{Learning-based adaptive control}

With the rapid advancement of highly parallel simulators~\cite{rudin2021learning,mittal2023orbit,xu2023omnidrones}, reinforcement learning (RL) methods are increasingly demonstrating powerful capabilities to achieve robust control, including quadrotors~\cite{zhang2023learning, huang2023datt} and legged robots~\cite{doi:10.1126/scirobotics.aau5872, kumar2021rma, long2023him, li2024reinforcement, radosavovic2024real}. Unlike classical model-based control, these learning-based methods often use model-free RL algorithms like PPO~\cite{schulman2017proximal} to train neural controllers in large-scale randomized environments. Koryakovskii et al.~\cite{modelplant2018ral} proposed to incorporate an RL-baesd policy to generate residual actions to compensate nonlinear model predictive control in model-mismatch cases. Rapid Motor Adaptation (RMA)~\cite{kumar2021rma} is a pure RL-based methods, which predicts environmental parameters in the latent space using state-action histories through a teacher-student training procedure. This approach has been extended to quadrotors~\cite{zhang2023learning}, bipedal robots~\cite{kumar2022adapting}, and humanoid robots~\cite{radosavovic2024real}. A follow-up study found that a dual-history architecture, leveraging both short-term and long-term input/output history, achieves superior performance with a straightforward one-stage training process~\cite{li2024reinforcement}. Conversely, DATT~\cite{huang2023datt} incorporates an \(\mathcal{L}_1\) adaptation law~\cite{l1ac} to estimate forces applied to quadrotors directly, which are then used as policy inputs. 
\revise{CAL~\cite{modelplant2018ral} integrates RL with nonlinear model predictive control (NMPC) by employing RL to generate residual actions that augment NMPC’s outputs. Although both DATT and CAL have proven effective with precise parameter tuning, they remain sensitive to handcrafted settings in their classical control components, limiting their generalizability to more complex systems.}

\section{Methods}
\label{sec:method}

In this section, we present a general framework that bridges model-based and RL-based control through a simplified predictive model, named NeuralIMC. We start by introducing the basics of Internal Model Control (IMC) and our extension to integrate it with RL-based neural control (\cref{sec:imc}). By leveraging the model-free RL algorithm PPO~\cite{schulman2017proximal}, we reduce the complexity and accuracy requirements typically associated with classical model-based control. Next, we detail how rigid transformation is used to predict the robot's body state, establishing a general approach to estimate disturbances and incorporate them into the control policy (\cref{sec:npm}). Finally, we explain the application of this framework to two specific robot morphologies: quadrotors and quadrupedal robots (\cref{sec:implementation}).

\subsection{Internal Model Control}
\label{sec:imc}
\begin{figure}[t] 
    \vspace{1.2mm}
    \centering
    \begin{subfigure}[b]{0.8\linewidth}
        \centering
        \includegraphics[width=0.9\linewidth]{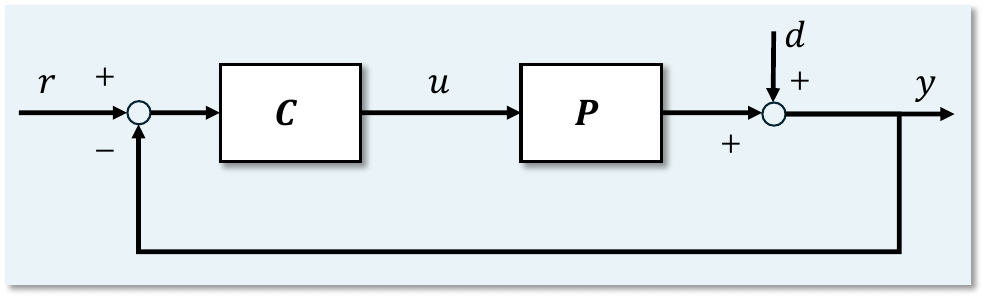}
        \caption{Classical Feedback Control}
        \label{fig:feedforward}
    \end{subfigure}
    \vspace{0.2cm}
    \begin{subfigure}[b]{0.8\linewidth}
        \centering
        \includegraphics[width=0.9\linewidth]{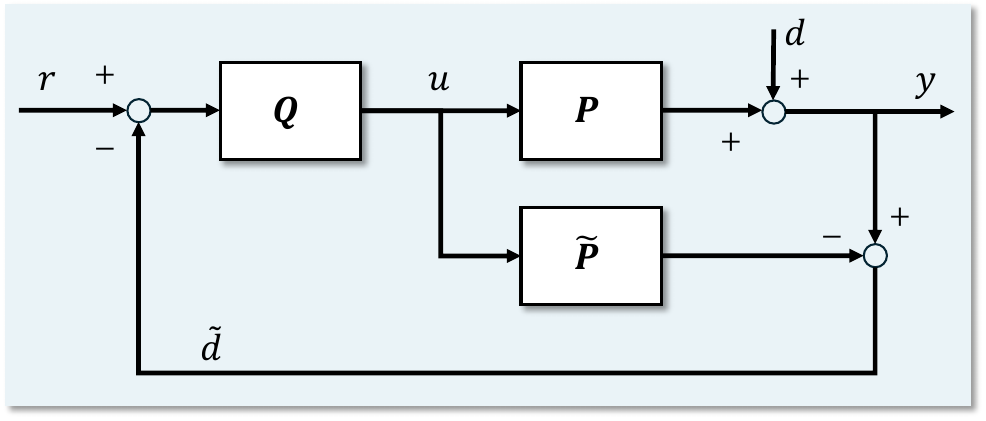}
        \caption{Internal Model Control}
        \label{fig:imc}
    \end{subfigure}
    \caption{Feedback structures of classical feedback control and Internal Model Control (IMC).}
    \label{fig:control_structure}
\end{figure}
Internal Model Control (IMC) is a robust control framework that outperforms classical feedback control in handling model uncertainties and disturbances. Unlike feedback control, which corrects errors in real time, IMC uses a predictive model to anticipate and compensate for disturbances. \cref{fig:control_structure} illustrates their structures, where $P$ and $\tilde{P}$ are the plant and internal models, $Q$ is the IMC controller, and $C$ is the classical feedback controller. The reference, disturbance, and output are denoted by $r$, $d$, and $y$, respectively. IMC computes the difference $\tilde{d} = \tilde{y} - y$ between predicted and actual responses, adjusting control inputs accordingly for enhanced robustness.

IMC is straightforward when the plant is easily modeled~\cite{rivera1986internal} but becomes challenging for nonlinear systems where approximations fail to capture true dynamics~\cite{henson1991internal,bin2020approximate}. This reliance on precise models limits its use in robotics, such as quadrotors with suspended payloads or legged robots on uneven terrain, where accurate system modeling is difficult. Our method integrates IMC with large-scale RL training to address these challenges. A simplified predictive model provides a rough yet stable estimate, while large-scale randomized simulations train a robust control policy that compensates for modeling inaccuracies. RL optimizes the policy with the predictive error feedback, revealing external disturbances and nonlinear dynamics, leading to more adaptive and effective control in real-world settings.

\subsection{Design of a Simplified Predictive Model}
\label{sec:npm}

The predictive model is key to our framework, enabling discrepancy detection and compensation. Its design faces two challenges: accurately estimating system response and closely aligning predictions with actual behavior. Precise estimation ensures the feedback loop reflects the true robot state, while alignment ensures reliability under varying conditions.

To streamline state estimation across diverse robots, we model only the rigid body state, excluding actuator- or limb-specific states. For robots equipped with monocular or binocular cameras and built-in IMUs, numerous visual(-inertial) odometry methods provide robust state estimations~\cite{qin2018vins,von2018direct,t265,NEURIPS2023_7ac484b0}, which can be used as the actual system response within the IMC structure. By treating the robot as a rigid body and using Euclidean transformation to estimate the next state, we significantly reduce the modeling complexity, making it more practical and efficient.

Here, we first present a general approach for formulating the predictive model and calculating the error, followed by specific adaptations for different robot morphologies in the next section. We assume that the model inputs include velocity or acceleration commands, and define the current body state as \(\mathbf{x}_k = [\mathbf{p}_k, \mathbf{q}_k, \mathbf{v}_k, \boldsymbol{\omega}_k] \in \mathbb{R}^{13}\), where \(\mathbf{p}_k\) is the position, \(\mathbf{q}_k\) is the quaternion (orientation), \(\mathbf{v}_k\) is the velocity, and \(\boldsymbol{\omega}_k\) is the angular velocity. The time interval between two steps is denoted as \(\Delta t\).

The next position and velocity of the body can be computed via Newton's law~\cite{newton1687principia}:
\begin{align}
    \label{eq:update_pos}
    \tilde{\mathbf{p}}_{k+1} &= \mathbf{p}_k + \mathbf{v}_k \Delta t + \frac{1}{2} \mathbf{a}_k \Delta t^2, \\
    \label{eq:update_vel}
    \tilde{\mathbf{v}}_{k+1} &= \mathbf{v}_k + \mathbf{a}_k \Delta t.
\end{align}
For rotation, we adopt a discrete update rule while handling singularities. First, we compute the change in angle \(\Delta \boldsymbol{\theta}\) with the angular velocity \(\boldsymbol{\omega}_k\) and angular acceleration \(\dot{\boldsymbol{\omega}}_k\):
\begin{align}
    \label{eq:update_dang}
    \Delta \boldsymbol{\theta} &= \boldsymbol{\omega}_k \Delta t + \frac{1}{2} \dot{\boldsymbol{\omega}}_k \Delta t^2.
\end{align}
Next, we follow \cite{grassia1998practical} to apply the exponential map to project the rotation from \(\mathbb{R}^3\) to \(SO(3)\), obtaining the corresponding quaternion \(\mathbf{q}_{\Delta \theta}\):
\begin{align}
\label{eq:update_dquat}
    \mathbf{q}_{\Delta \theta} &= \begin{cases}
        \left[\cos\left(\frac{\|\Delta \boldsymbol{\theta}\|}{2}\right), \frac{\sin\left(\frac{\|\Delta \boldsymbol{\theta}\|}{2}\right)}{\|\Delta \boldsymbol{\theta}\|} \cdot \Delta \boldsymbol{\theta}\right], \|\Delta \boldsymbol{\theta}\|\leq \sqrt[4]{\epsilon_{\text{machine}}} \\
        \left[\cos\left(\frac{\|\Delta \boldsymbol{\theta}\|}{2}\right), \left(\frac{1}{2}+\frac{\|\Delta \boldsymbol{\theta}\|^2}{48}\right) \cdot \Delta \boldsymbol{\theta}\right], \text{otherwise}
    \end{cases}
\end{align}
Finally, we can predict the next quaternion and angular velocity as follows, where \(\otimes\) denotes quaternion multiplication:
\begin{align}
    \label{eq:update_quat}
    \tilde{\mathbf{q}}_{k+1} &= \mathbf{q}_k \otimes \mathbf{q}_{\Delta \theta} \\
    \label{eq:update_omega}
    \tilde{\mathbf{\omega}}_{k+1} &= \mathbf{\omega}_k + \dot{\mathbf{\omega}}_k \Delta t
\end{align}
We have established an explicit method to predict the next body state, assuming the required motion can be represented by commands at the acceleration or velocity level. Then, given the predicted body state \(\tilde{\mathbf{x}}_{k+1} = \left[\tilde{\mathbf{p}}_{k+1}, \tilde{\mathbf{q}}_{k+1}, \tilde{\mathbf{v}}_{k+1}, \tilde{\mathbf{\omega}}_{k+1}\right]\) and the actual body state \(\mathbf{x}_{k+1}\), we compute the predictive error \(\tilde{\mathbf{d}}_{k+1}\), which is then fed into the policy to provide effective feedback signals.
\begin{align}
    \tilde{\mathbf{d}}_{k+1} = \left[\Delta \tilde{\mathbf{p}}_{k+1}, 1 - \cos{\tilde{\Theta}}, \Delta \tilde{\mathbf{v}}_{k+1}, \Delta \tilde{\mathbf{\omega}}_{k+1}\right]
\end{align}
Here, \(\Delta \tilde{\mathbf{p}}_{k+1}\), \(\Delta \tilde{\mathbf{v}}_{k+1}\), and \(\Delta \tilde{\mathbf{\omega}}_{k+1}\) represent the differences in position, velocity, and angular velocity, respectively. We use the cosine distance to quantify the difference between orientations, where \(\tilde{\Theta}\) is the angle difference between the two quaternions.

\subsection{Specifications for Common Robots}
\label{sec:implementation}
In this section, we further explain how to derive velocity or acceleration commands for various robots, including quadrotors and quadrupedal robots. We demonstrate how these commands align with existing control methods, ensuring seamless integration into the policy learning process. 

\subsubsection{Quadrotors}
Traditional quadrotor controllers use cascaded structures~\cite{px4controller}, allowing flexible integration of neural controllers. Recent studies show that using collective thrust and bodyrate (CTBR) as the action space enhances resilience to dynamics mismatches, improves domain transfer, and boosts agility~\cite{kaufmann2022benchmark, huang2023datt,kaufmann2023champion}. This design aligns with our framework: the policy’s bodyrate command updates rotation, while collective thrust determines linear acceleration. Consequently, predictive errors, computed from prior policy output and successive body states, integrate directly into the current policy input without modifying the learning algorithm.

\subsubsection{Quadrupedal robots}
Unlike quadrotors, legged robots, particularly quadrupeds and bipeds, are well-suited for RL-based control due to their high-dimensional actuation and terrain interactions~\cite{li2023learning, doi:10.1126/scirobotics.aau5872, kumar2021rma, li2024reinforcement, li2022bridging}. Neural controllers typically generate actuator targets to achieve desired body velocities (\(v_x, v_y, \omega_z\)). To integrate our predictive model into the control loop,, we use reference velocity commands as model inputs, deriving linear and angular accelerations by differentiating consecutive commands. Setting other accelerations to zero ensures seamless integration with existing policies. This method is compatible with other algorithms via a plug-and-play feedback loop augmenting policy inputs.

\section{Experimental Results}
\label{sec:exp}

To evaluate our framework's performance and versatility, we conduct experiments on quadrotors and quadrupedal robots, comparing against state-of-the-art methods in each domain. Then, we perform ablation studies on the feedback structure and noise tolerance of the simplified predictive model. Finally, we present sim-to-real deployment results on a quadrotor with two rope-suspended payloads, showcasing its effectiveness in enhancing system robustness.

\begin{table}[t]
    \centering
    \vspace{1.2mm}
    \caption{Parameter ranges for dynamics randomization on quadrotor. $K_p^0=[37, 37, 11]$}.
    \vspace{1mm}
    \begin{tabular}{c|cc}
        \toprule
        Parameter Name & Train Range & Eval Range \\
        \midrule
        Mass ($M$) & [0.142, 0.950] & [0.114, 1.140] \\
        Arm Length ($L$) & [0.046, 0.200] & [0.037, 0.240] \\
        Width-to-$L$ Ratio ($\alpha_w$) & [1.0, 1.414] & [0.852, 1.732] \\
        Height-to-$L$ Ratio ($\alpha_h$) & [0.577, 1.0] & [0.268, 1.414] \\
        Motor Force Constant ($K_f$) & [1.15e-7, 7.64e-6] & [9.16e-8, 9.17e-6] \\
        Max Thrust-to-Weight Ratio & [2.0, 3.5] & [1.6, 4.2] \\
        Motor Drag Constant ($\kappa$) & [0.0041, 0.0168] & [0.0033, 0.0201] \\
        Motor Time Constant ($\tau_c$) & [0.3, 0.5] & [0.24, 0.6] \\
        P Gains ($K_p$) & [-0.3,0.3] $\times K_p^0$ & [-0.5, 0.5] $\times K_p^0$ \\
        \bottomrule
    \end{tabular}
    \label{tab:quadrotor_random}
\end{table}

\begin{table*}[t]
    \vspace{1.2mm}
    \caption{Simulation results for tracking errors (unit: m) with quadrotors, averaged over five random seeds. \textbf{Bold values} denote the best results. \colorbox{cyan!10}{Blue cells} indicate performance under random \revise{external force disturbances}, with percentages in brackets representing the relative change in error compared to static conditions. T and S refer to the teacher and student models for RMA, respectively, while DATT with GT indicates the use of ground-truth disturbance inputs instead of $\mathcal{L}_1$ estimates.}
    \centering
    \begin{tabular}{c|c|cc|cc|cc|cc}
        \toprule
        \multirow{2}{*}{Config} &\multirow{2}{*}{Method} &\multicolumn{4}{c|}{Smooth Trajectory}  &\multicolumn{4}{c}{Infeasible Trajectory}\\
        \cline{3-10}
        & &\multicolumn{2}{c|}{Circle} &\multicolumn{2}{c|}{Chained Poly} &\multicolumn{2}{c|}{5-Point Star} &\multicolumn{2}{c}{Zigzag}\\
        \midrule
        \multirow{9}{*}{Train} 
        &\revise{$\mathcal{L}_1$-PID} &\revise{0.214} &\cellcolor{cyan!10}\revise{0.237 (+10.74\%)} &\revise{0.043} &\cellcolor{cyan!10}\revise{0.054 (+25.58\%)} &\multicolumn{2}{c|}{\revise{\textit{crashed}}}  &\multicolumn{2}{c}{\revise{\textit{crashed}}}\\
        &\revise{$\mathcal{L}_1$-MPPI} &\revise{0.202} &\cellcolor{cyan!10}\revise{0.215 (+6.44\%)} &\revise{0.092} &\cellcolor{cyan!10}\revise{0.117 (+27.17\%)} &\revise{0.193} &\cellcolor{cyan!10}\revise{0.216 (+11.91\%)} &\revise{0.169} &\cellcolor{cyan!10}\revise{0.186 (+10.06\%)}  \\
        &\revise{CAL} &0.087 &\cellcolor{cyan!10}0.157 (+80.46\%) &0.056 &\cellcolor{cyan!10}0.118 (+110.71\%) &0.103 &\cellcolor{cyan!10}0.175 (+69.90\%) &0.105 &\cellcolor{cyan!10}0.164 (+56.19\%) \\
        &PPO & 0.102 &\cellcolor{cyan!10}0.169 (+65.69\%) &0.065 &\cellcolor{cyan!10}0.117 (+80.00\%) &0.097 &\cellcolor{cyan!10}0.172 (+77.32\%) &0.095 &\cellcolor{cyan!10}0.154 (+62.11\%)\\
        &RMA (T) &\textbf{0.071} &\cellcolor{cyan!10}\textbf{0.080} (+12.68\%) &0.043 &\cellcolor{cyan!10}0.047 ({+9.30\%}) &0.087 &\cellcolor{cyan!10}0.094 (\textbf{+8.05\%}) &0.074 &\cellcolor{cyan!10}0.082 ({+10.81\%})\\
        &RMA (S) &0.084 &\cellcolor{cyan!10}0.110 (+30.95\%) &0.048 &\cellcolor{cyan!10}0.064 (+33.33\%) &0.116 &\cellcolor{cyan!10}0.157 (+35.34\%) &0.112 &\cellcolor{cyan!10}0.142 (+26.79\%) \\
        &DATT ($\mathcal{L}_1$) & 0.084 &\cellcolor{cyan!10}0.128 (+52.38\%) &0.051 &\cellcolor{cyan!10}0.076 (+49.02\%) &0.089 &\cellcolor{cyan!10}0.163 (+83.15\%) &0.077 &\cellcolor{cyan!10}0.138 (+79.22\%)\\
        &DATT (GT) &0.084 &\cellcolor{cyan!10}0.093 (\textbf{+10.71\%}) &0.052 &\cellcolor{cyan!10}0.055 (\textbf{+5.77\%}) &0.089 &\cellcolor{cyan!10}0.098 (+10.11\%) &0.077 &\cellcolor{cyan!10}0.084 (\textbf{+9.09\%}) \\
        &{Ours} & 0.072 &\cellcolor{cyan!10}0.083 (+15.28\%) &\textbf{0.032} &\cellcolor{cyan!10}\textbf{0.038} (+18.75\%) &\textbf{0.065} &\cellcolor{cyan!10}\textbf{0.073} (+12.31\%) &\textbf{0.057} &\cellcolor{cyan!10}\textbf{0.064} (+12.28\%) \\
        \midrule
        \multirow{9}{*}{Eval} 
        &\revise{$\mathcal{L}_1$-PID} &\revise{0.372} &\cellcolor{cyan!10}\revise{0.416 (+11.82\%)} &\revise{0.061} &\cellcolor{cyan!10}\revise{0.107 (+75.41\%)} &\multicolumn{2}{c|}{\revise{\textit{crashed}}}  &\multicolumn{2}{c}{\revise{\textit{crashed}}}\\
        &\revise{$\mathcal{L}_1$-MPPI} &\revise{0.212} &\cellcolor{cyan!10}\revise{0.234 (+10.38\%)} &\revise{0.097} &\cellcolor{cyan!10}\revise{0.132 (+36.08\%)} &\revise{0.207} &\cellcolor{cyan!10}\revise{0.253 (+22.22\%)} &\revise{0.176} &\cellcolor{cyan!10}\revise{0.218 (+23.86\%)} \\
        &\revise{CAL} &0.224 &\cellcolor{cyan!10}0.321 (+43.30\%) &0.171 &\cellcolor{cyan!10}0.251 (+46.78\%) &0.262 &\cellcolor{cyan!10}0.364 (+38.93\%) &0.265 &\cellcolor{cyan!10}0.358 (+35.09\%)\\
        &PPO & 0.265 &\cellcolor{cyan!10}0.351 (+32.45\%) &0.183 &\cellcolor{cyan!10}0.273 (+49.18\%) &0.257 &\cellcolor{cyan!10}0.374 (+45.53\%) &0.265 &\cellcolor{cyan!10}0.367 (+38.49\%)\\
        &RMA (T) &0.117 &\cellcolor{cyan!10}0.147 (+25.64\%) &0.072 &\cellcolor{cyan!10}0.097 (+34.72\%) &0.151 &\cellcolor{cyan!10}0.192 (+27.15\%) &0.145 &\cellcolor{cyan!10}0.187 (+28.97\%)\\ 
        &RMA (S) &0.280 &\cellcolor{cyan!10}0.360 (+28.57\%) &0.174 &\cellcolor{cyan!10}0.243 (+39.66\%) &0.358 &\cellcolor{cyan!10}0.424 (+18.44\%) &0.349 &\cellcolor{cyan!10}0.417 (\textbf{+19.48\%})\\
        &DATT ($\mathcal{L}_1$) & 0.137 &\cellcolor{cyan!10}0.523 (+281.75\%) &0.094 &\cellcolor{cyan!10}0.354 (+276.60\%) &0.169 &\cellcolor{cyan!10}0.585 (+246.15\%) &0.160 &\cellcolor{cyan!10}0.559 (+249.38\%)\\
        & DATT (GT) &0.139 &\cellcolor{cyan!10}0.167 (+20.14\%) &0.094 &\cellcolor{cyan!10}0.119 (\textbf{+26.60\%}) &0.178 &\cellcolor{cyan!10}0.197 (\textbf{+10.67\%}) &0.156 &\cellcolor{cyan!10}0.189 (+21.15\%) \\
        &{Ours} & \textbf{0.113} &\cellcolor{cyan!10}\textbf{0.135} (\textbf{+19.47\%}) &\textbf{0.054} &\cellcolor{cyan!10}\textbf{0.073} (+35.19\%) &\textbf{0.103} &\cellcolor{cyan!10}\textbf{0.134} (+30.10\%) &\textbf{0.101} &\cellcolor{cyan!10}\textbf{0.132} (+30.69\%)\\
        \bottomrule
    \end{tabular}
    \vspace{-2mm}
    \label{tab:main_quadrotor}
\end{table*}

\subsection{Experiment Settings}

\subsubsection{Quadrotors}

For quadrotors, controllers are trained using a customized simulator based on the dynamics shown in \cite{faessler2016thrust}, where the collective thrust and bodyrate commands are followed by a PD controller. The fourth-order Runge-Kutta method~\cite{suli2003introduction} updates quadrotor states at each simulation step with a update frequency at 200 Hz, and the controller's execution frequency is 50 Hz. Following Huang et al.~\cite{huang2023datt}, we train controllers on two types of reference trajectories, including smooth chained polynomial trajectories and infeasible zigzag trajectories. External disturbances are applied as randomized force perturbations within \([-3.3 \, \text{m/s}^2, 3.3 \, \text{m/s}^2]\), representing Brownian motion variations as in \cite{huang2023datt}. Dynamical parameters are randomized at the start of each episode for training, and unseen parameters are used for evaluation to test robustness. The detailed randomization range is shown in \cref{tab:quadrotor_random}. We train and evaluate each method using five random seeds, reporting the average performance across 1,024 parallel trials. For evaluation, we test controllers trained with smooth trajectories on two types of randomly generated smooth paths: circle and chained polynomial, while controllers trained on infeasible paths are tested on 5-point star and zigzag trajectories.

\subsubsection{Quadrupedal robots}

Following Long et al.~\cite{long2023him}, we train controllers for the Unitree Aliengo using Isaac Gym~\cite{rudin2021learning} with 4096 parallel environments and a 100-step rollout. The training curriculum includes progressively challenging terrains and reference commands. Controllers are trained on terrains such as pyramid slopes, rough pyramid slopes, stairs, and flats with discrete obstacles, with distributions of [0.1, 0.2, 0.6, 0.1]. We evaluate methods by tracking x-y linear velocity or yaw rate commands under two conditions: no disturbances and random force pushes. The x-y linear velocity command is sampled from \([-1.0, 1.0] \, \text{m/s}\), and the yaw rate command from \([-1.0, 1.0] \, \text{rad/s}\). 
During linear velocity tracking evaluations, the angular velocity target is set to zero, and vice versa for angular velocity tracking. For linear velocity tracking, the forward and lateral velocities are uniformly sampled from \([-1 \, \text{m/s}, 1 \, \text{m/s}]\). For angular velocity tracking tasks, the yaw rate is uniformly sampled from \([-1 \, \text{rad/s}, 1 \, \text{rad/s}]\). 
The tracking error is averaged over five random seeds, each with 4096 parallel environments. To validate robustness, we emulate an impulse push by setting a randomized base velocity to the robot every 16 steps, uniformly sampled within the maximum range. The maximum impulse velocity during training is 1 m/s, increased to 1.5 m/s for evaluation. 

\subsection{Experiments on Quadrotors}
\label{sec:exp_quadrotor}
We compare our method for quadrotors against \revise{two classical control methods ($\mathcal{L}_1$-PID and $\mathcal{L}_1$-MPPI) and four learning-based baselines (CAL~\cite{modelplant2018ral}, PPO~\cite{schulman2017proximal}, RMA~\cite{kumar2021rma}, DATT~\cite{huang2023datt}}), using two variants each for RMA and DATT to enable deeper analyses. All methods use the same PPO training parameters. For \revise{CAL~\cite{modelplant2018ral}, a sum combination of MPPI and RL, MPPI serves as a sampling-based non-linear MPC method, with the RL policy compensating the MPPI output.} We set the sampling number to 4096 and the rollout horizon to 20 in MPPI's sampling procedure.
PPO employs a three-layer MLP policy and a temporal convolutional network (TCN) to process future waypoints in the reference trajectory, which is inherited by all other methods. In addition, RMA adopts a teacher-student training framework, with RMA(T) as the teacher model and RMA(S) as the student. RMA(T) integrates dynamic parameters and ground-truth force disturbances, while RMA(S) uses a TCN to approximate the teacher’s latent state from state-action histories. In contrast, DATT utilizes an $\mathcal{L}_1$ disturbance estimator and tracking error inputs to improve trajectory-tracking robustness. To separate the $\mathcal{L}_1$ estimator’s effect, we also include a privileged variant, DATT(GT), which directly uses ground-truth force disturbances in place of the estimator.

\final{As shown in \cref{tab:main_quadrotor}, our method consistently outperforms others across nearly all tasks, particularly under disturbances and out-of-distribution (OOD) system dynamics.}

\final{Among estimator-based baselines—both explicit (e.g., \(\mathcal{L}_1\) estimator in \(\mathcal{L}_1\)-PID, \(\mathcal{L}_1\)-MPPI, DATT) and implicit (e.g., RMA’s neural encoder)—DATT is the most resilient, benefiting from its explicit \(\mathcal{L}_1\) estimator coupled with an RL-based controller. Nevertheless, all estimator-based methods degrade sharply under our extreme out-of-distribution scenarios—where internal dynamics shift or sudden external forces occur—because their estimators become unreliable and control robustness is compromised. Estimator-free approaches such as CAL and PPO also fail to manage these varied disturbances. In contrast, our method computes the predictive error between expected and observed body transformations, directly signaling state drift and disturbance magnitude. A policy trained on this error maintains strong performance across both nominal and unexpected conditions.}

\final{We also acknowledge that limited hyperparameter sweeps in our experiments may contribute to the baselines’ degraded performance. To ensure a fair comparison, we include privileged variants (DATT (GT) and RMA (T)) that receive ground-truth states rather than estimates, representing an upper tuning bound. Even against these privileged methods, our predictive error-driven controller achieves superior and more stable control, supporting our hypothesis that explicit predictive error feedback offers a simple, generalizable path to robust performance.}

\begin{table*}[t]
    \vspace{1.2mm}
    \caption{Simulation results on quadrupedal robots, averaged over five random seeds. The tracking error for linear velocity and angular velocity are computed as \(\|v_{x,y} - v_{x,y}^{\text{target}}\|_2\) (unit: m/s) and \(\|\omega_{\text{yaw}} - \omega_{\text{yaw}}^{\text{target}}\|_2\) (unit: rad/s), respectively. The best results for each case are highlighted \textbf{in bold}.}
    \centering
    \begin{tabular}{c|c|cc|cc|cc|cc}
        \toprule
        \multirow{2}{*}{Disturbance}&\multirow{2}{*}{Method} &\multicolumn{2}{c|}{Slopes} &\multicolumn{2}{c|}{Rough Slopes} &\multicolumn{2}{c|}{Stairs} &\multicolumn{2}{c}{Discrete Obstacles}\\
        & &Linear &Angular &Linear &Angular &Linear &Angular &Linear &Angular \\
        \midrule
        \midrule
        \multirow{4}{*}{None} &PPO & 0.132 & 0.085 & 0.143 & 0.087 & 0.200 & 0.085 & 0.163 & 0.085 \\
&\revise{RMA} & \revise{0.084} & \revise{0.050} & \revise{0.086} & \revise{0.051} & \revise{0.127} & \revise{0.050} & \revise{0.110} & \revise{0.050}  \\
&HIM & 0.100 & 0.067 & 0.103 & 0.068 & 0.181 & 0.067 & 0.147 & 0.067 \\
&{Ours (w/ HIM)} & 0.077 & 0.052 & 0.079 & 0.053 & 0.126 & 0.053 & 0.105 & 0.052 \\
&{Ours} & \textbf{0.071} & \textbf{0.048} & \textbf{0.073} & \textbf{0.049} & \textbf{0.101} & \textbf{0.048} & \textbf{0.085} & \textbf{0.048} \\
        \midrule
        \multirow{4}{*}{Impulse} &PPO & 0.151 & 0.101 & 0.162 & 0.103 & 0.216 & 0.101 & 0.178 & 0.101 \\
&\revise{RMA} & \revise{0.096} & \revise{0.059} & \revise{0.098} & \revise{0.060} & \revise{0.137} & \revise{0.060} & \revise{0.120} & \revise{0.059} \\
&HIM & 0.117 & 0.080 & 0.120 & 0.081 & 0.186 & 0.080 & 0.157 & 0.080 \\
&{Ours (w/ HIM)} & 0.091 & 0.061 & 0.093 & 0.061 & 0.139 & 0.061 & 0.117 & 0.061 \\
&{Ours} & \textbf{0.086} & \textbf{0.058} & \textbf{0.087} & \textbf{0.059} & \textbf{0.115} & \textbf{0.059} & \textbf{0.099} & \textbf{0.058} \\
        \bottomrule
    \end{tabular}
    \vspace{-2mm}
    \label{tab:main_quadruped}
\end{table*}

\subsection{Experiments on Quadrupedal Robots}
\label{sec:exp_quadruped}
\revise{For quadrupedal robots, we compare our methods with PPO~\cite{schulman2017proximal}, RMA~\cite{kumar2021rma} and HIM~\cite{long2023him}. PPO and RMA serve as standard and common baselines in continuous control for quadrupeds, while HIM leverages unsupervised contrastive learning to train a predictive encoder for the next latent state and body velocity. Here, we directly use the teacher-policy of RMA to set a strong baseline.} Our method enhances the PPO policy by integrating predictive error feedback. Additionally, we evaluate a variant of our method, adding predictive error feedback to both the inputs of the HIM policy and its predictive encoder during training and evaluation.

\revise{As shown in \cref{tab:main_quadruped}, our method outperforms HIM by 26–44\% across various test conditions and matches or surpasses the privileged RMA.}
The combination of our framework with HIM also shows improvements over the original HIM. Interestingly, the predictive encoder in HIM shows a slight degradation compared to standard PPO when combined with our NeuralIMC structure. We hypothesize that this is because the predictive error provides sufficient signal for robust control, rendering the additional impact of predictive encoding negligible.
This suggests that predictive error feedback alone is sufficiently informative for enhancing control robustness, simplifying the neural control pipeline, and facilitating integration with other learning algorithms.

\subsection{Ablation Studies}
\label{sec:ablation}
\revise{In this section, we conduct three ablation studies to analyze our framework. Inspired by \cite{li2024reinforcement}, we first examine whether incorporating short- and long-horizon input/output histories improves performance. Next, we introduce noise into the predictive model's input and output to assess its robustness and the impact of prediction accuracy on control. Finally, we compare our method using the predictive model versus a full-dynamics model to evaluate the effect of model simplification. All experiments involve quadrotors tracking zigzag trajectories across three random seeds.}

\begin{figure}[t] 
    \centering
    \begin{subfigure}[b]{0.8\linewidth}
        \centering
        \includegraphics[width=\linewidth]{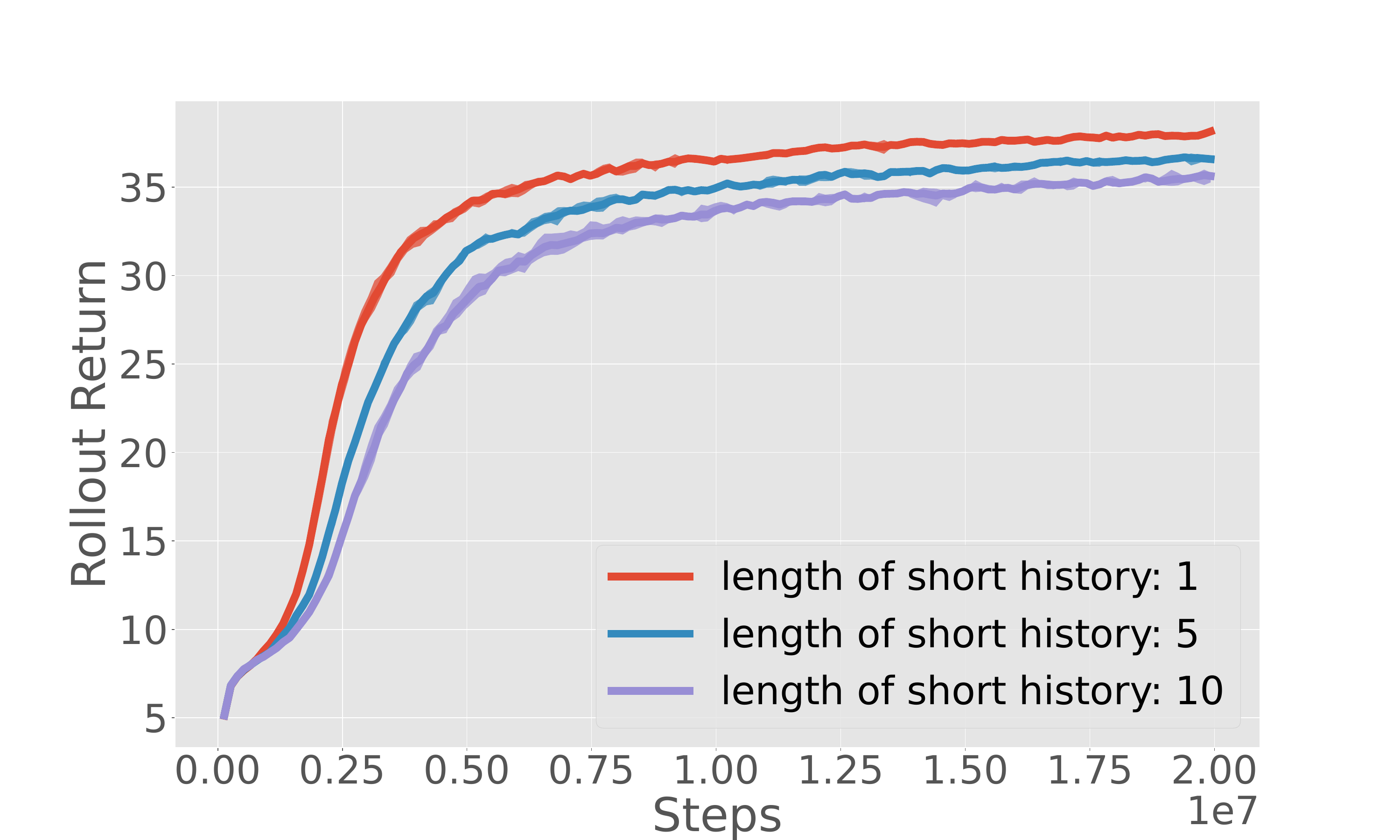}
        \caption{Varying short-history length.}
        \label{fig:short-h}
    \end{subfigure}
    \begin{subfigure}[b]{0.8\linewidth}
        \centering
        \includegraphics[width=\linewidth]{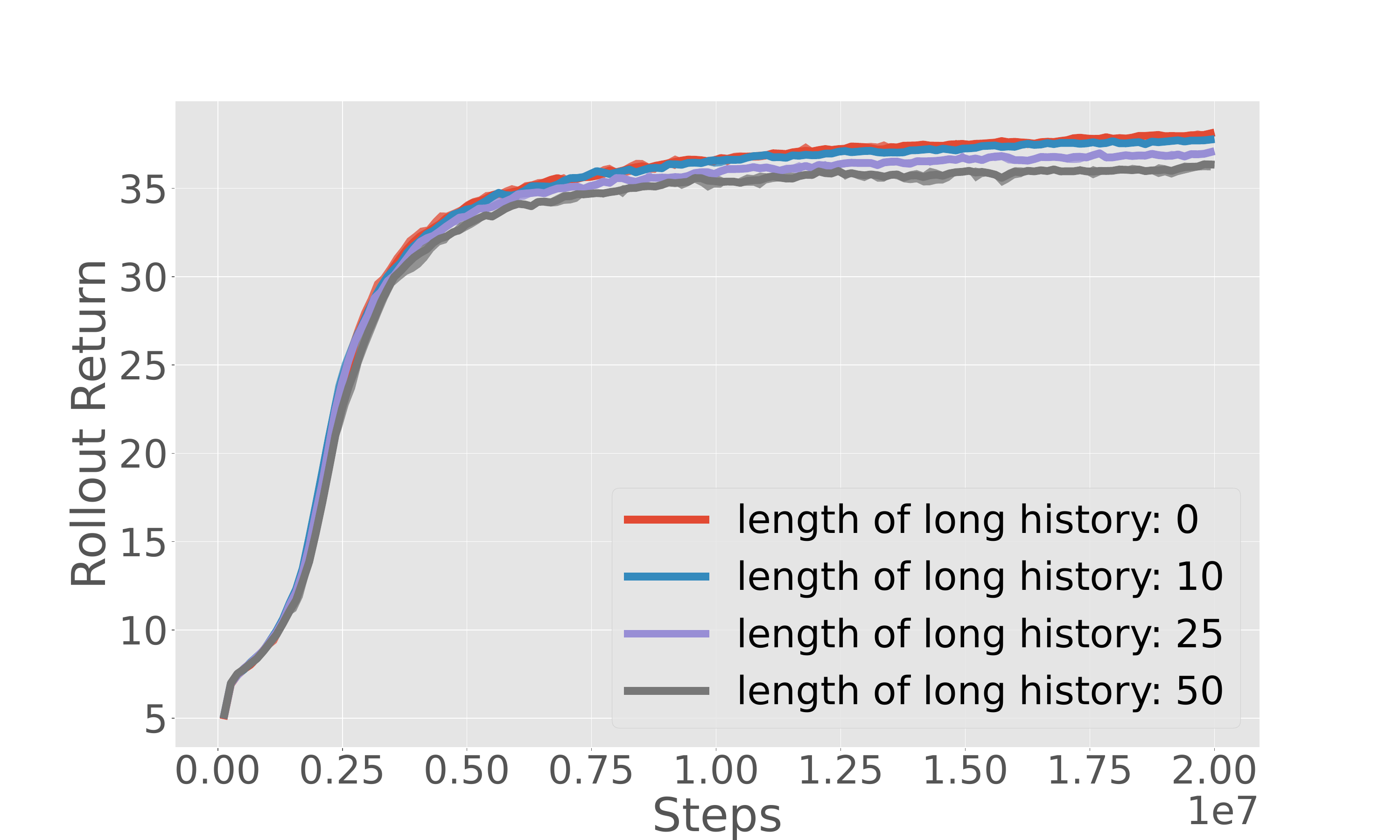}
        \caption{Varying long-history length.}
        \label{fig:long-h}
    \end{subfigure}
    \caption{Ablation study on the feedback structure with different history lengths.}
    \vspace{-3mm}
    \label{fig:ablation_history}
\end{figure}

\begin{table}[t]
    \centering
    \caption{\revise{Evaluation of the internal predictive model's noise tolerance, tested on quadrotors tracking zigzag trajectories.}}
    \scalebox{0.9}{
    \begin{tabular}{c|c|c|c|c|c}
        \toprule
        &\multirow{2}{*}{Noise $\sigma_n$}&\multicolumn{2}{c|}{Train} &\multicolumn{2}{c}{Eval}  \\
        \cline{3-6}
        &&static &disturbed &static &disturbed \\
        \midrule
        &0. &0.057 &\cellcolor{cyan!10}0.064 (+12.28\%) &0.101 &\cellcolor{cyan!10}0.119 (+17.82\%) \\
        \midrule
        \multirow{4}{*}{\makecell{input\\noise}} &0.02 &0.063 &\cellcolor{cyan!10}0.069 (+9.52\%) &0.092 &\cellcolor{cyan!10}0.117 (+27.17\%) \\
        &0.04 &0.070 &\cellcolor{cyan!10}0.094 (+34.28\%) &0.100 &\cellcolor{cyan!10}0.154 (+54.00\%) \\
        &0.08 &0.071 &\cellcolor{cyan!10}0.121 (+70.42\%) &0.134 &\cellcolor{cyan!10}0.202 (+50.74\%) \\
        &0.16 &0.082 &\cellcolor{cyan!10}0.135 (+64.63\%) &0.152 &\cellcolor{cyan!10}0.231 (+51.97\%) \\
        \midrule
        \multirow{4}{*}{\makecell{output\\noise}} &0.02 &0.063 &\cellcolor{cyan!10}0.071 (+12.69\%) &0.076 &\cellcolor{cyan!10}0.132 (+73.68\%) \\
        &0.04 &0.073 &\cellcolor{cyan!10}0.097 (+32.87\%) &0.102 &\cellcolor{cyan!10}0.161 (+57.84\%) \\
        &0.08 &0.074 &\cellcolor{cyan!10}0.121 (+63.51\%) &0.118 &\cellcolor{cyan!10}0.179 (+51.69\%) \\
        &0.16 &0.076 &\cellcolor{cyan!10}0.130 (+71.05\%) &0.138 &\cellcolor{cyan!10}0.222 (+60.86\%) \\
        \bottomrule
    \end{tabular}}
    \label{tab:noise}
\end{table}

\begin{figure*}[t]
    \centering
    \vspace{1.2mm}
    \begin{subfigure}[b]{0.3\linewidth}
        \centering
        \includegraphics[width=\linewidth]{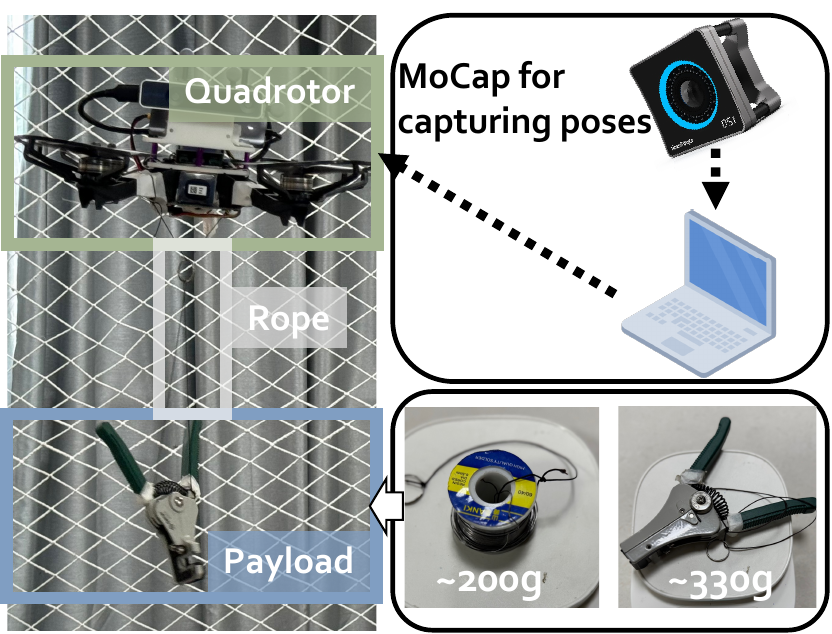}
        \caption{Quadrotor system for deployment.}
        \label{fig:system}
    \end{subfigure}
    \begin{subfigure}[b]{0.33\linewidth}
        \centering
        \includegraphics[width=\linewidth]{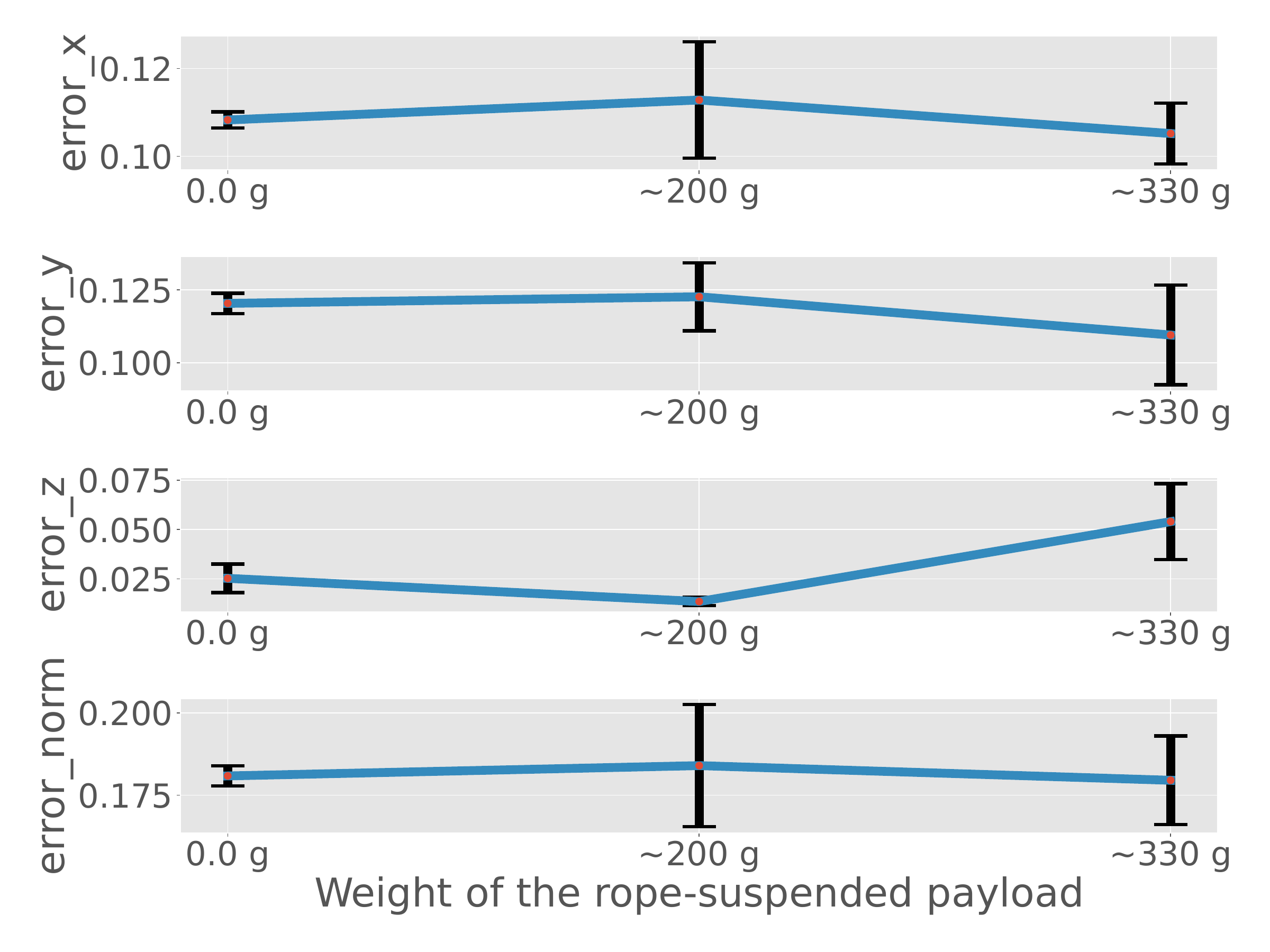}
        \caption{Performance across different disturbances.}
        \label{fig:payload}
    \end{subfigure}
    \begin{subfigure}[b]{0.33\linewidth}
        \centering
        \includegraphics[width=\linewidth]{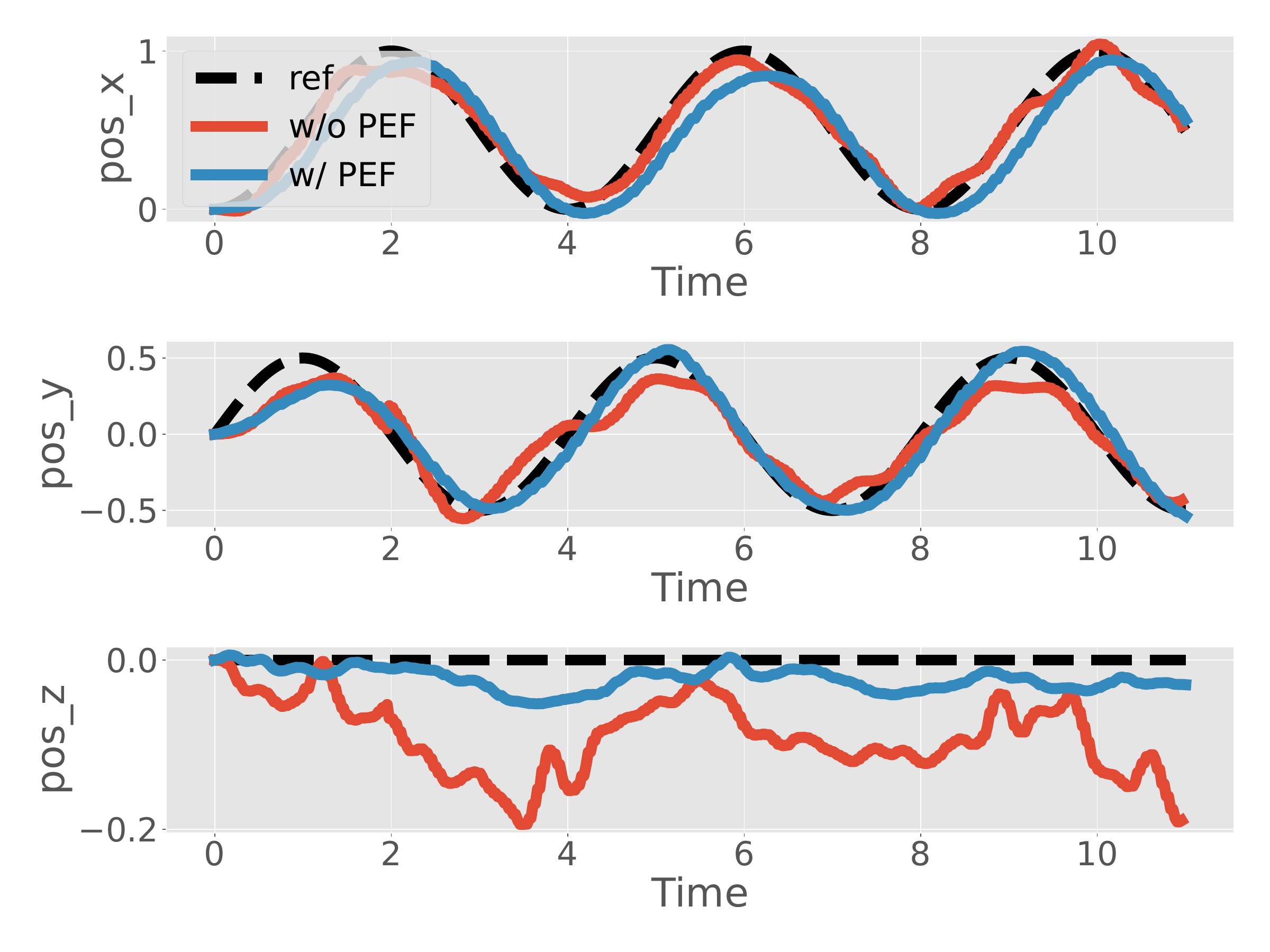}
        \caption{Effect of the predictive error feedback.}
        \label{fig:real_traj}
    \end{subfigure}
    \caption{Real-world experiments on a quadrotor with two different rope-suspending payloads.}
    \label{fig:real}
\end{figure*}

\subsubsection{History structure}
We evaluate feedback structures incorporating historical records. In short-history experiments, we concatenate recent input/output pairs, including predictive errors, into a single input vector. For long-history experiments, we retain the latest state-action pair while encoding additional history using a TCN model~\cite{kumar2021rma,li2024reinforcement}, concatenating its latent output with the policy input.

As shown in \cref{fig:ablation_history}, using longer history—either short- or long-term—does not improve performance over immediate feedback (i.e., the latest one-step history). While differing from \cite{li2024reinforcement}, this aligns with $\mathcal{L}_1$ adaptive control~\cite{l1ac} and robust control methods~\cite{nguyen2018model,rivera1986internal}, where immediate error feedback suffices for accuracy and robustness. This simpler structure also reduces implementation complexity and mitigates cumulative errors in real robots.

\subsubsection{Noise tolerance}
\label{sec:ab_noise}
To assess the impact of the internal predictive model’s accuracy on control performance, we introduce varying levels of noise into the model’s \revise{state input or predictions}. \revise{Specifically, we introduce zero-mean Gaussian noise with varying variances to the state vector, including position, velocity, quaternion, and angular velocity. For the quaternion, we treat the sampled Gaussian noise as a delta Euler angle, convert it to a delta quaternion, and apply it to rotate the original orientation.} We train and evaluate multiple policies using a range of noisy predictive models for quadrotor tracking on zigzag trajectories, with the results presented in \cref{tab:noise}.
 
\revise{The results show that our method maintains competitive performance under mild noise, whether applied to model inputs or responses. However, as noise levels increase, the accuracy of the predictive model becomes increasingly crucial}, particularly in scenarios with external disturbances or out-of-distribution system dynamics. Across different training and evaluation ranges, performance declines as noise intensifies, with a similar trend observed under external force disturbances. This observation highlights an advantage of our explicit predictive model over other learning-based approaches: it provides a stable baseline estimation, albeit approximate. Although modeling noise can affect feedback reliability, the rigid-body transformation adheres to fundamental motion principles and generalizes across robotic systems. While this model simplifies nonlinear complexities, the RL part in our framework compensates effectively within an informative closed-loop system.

\subsubsection{\revise{Model simplification}}
\label{sec:ab_model}
\begin{table}[t]
    \centering
    \caption{\revise{Evaluation of the internal predictive model's complexity, tested on quadrotors tracking zigzag trajectories.}}
    \scalebox{1.0}{
    \begin{tabular}{c|c|c|c|c}
        \toprule
        \multirow{2}{*}{Model}&\multicolumn{2}{c|}{Train} &\multicolumn{2}{c}{Eval}  \\
        \cline{2-5}
        &static &disturbed &static &disturbed \\
        \midrule
        Simplified &0.057 &\cellcolor{cyan!10}0.064 (+12.28\%) &0.101 &\cellcolor{cyan!10}0.119 (+17.82\%) \\
        Full &0.061 &\cellcolor{cyan!10}0.067 (+9.84\%) &0.095 &\cellcolor{cyan!10}0.105 (+10.53\%) \\
        \bottomrule
    \end{tabular}}
    \label{tab:model}
\end{table}

\revise{We investigate the impact of predictive model simplification on performance. A quadrotor serves as our testbed due to its well-studied dynamics model~\cite{faessler2016thrust}. We compare our approach with a variant that uses the full-dynamics model for tracking zigzag trajectories. As shown in \cref{tab:model}, the results indicate that despite the simplification, performance degradation is marginal. Notably, this simplification improves generalization across different morphologies, making it applicable to complex systems that are difficult to model.} 

\subsection{Real-world Deployment}

To further validate the effectiveness of our framework, we conducted a real-world experiment on a quadrotor, deploying our method to assess its robustness under various disturbances.
The quadrotor, equipped with an Nvidia Jetson Xavier NX~\cite{jetson_xavier_nx} and a PX4 flight control unit (FCU)~\cite{px4fcu}, interfaces with a motion capture system for real-time pose tracking. The whole neural control framework runs on the embedded computer. The control policy generates CTBR commands, transmitted to the FCU via MAVROS~\cite{mavros} and converted to motor velocities. The quadrotor weighs approximately 1 kg, and we evaluated its tracking performance across five trials under three disturbance conditions: no disturbance, a 200g payload, and a 330g payload, each attached via a rope to increase instability and nonlinearity. The deployment system setup is shown in \cref{fig:system}.

We present the tracking performance for each axis, along with the normalized error across all axes, in \cref{fig:payload}. The lines represent the average tracking errors under different conditions, with black intervals indicating variance. Our results show that the framework maintains relatively stable performance across various payloads. To illustrate the impact of predictive error feedback, we include a detailed example trajectory for the 330g payload as an ablation study in a real-world scenario (\cref{fig:real_traj}). The z-axis trajectory demonstrates that, without predictive error feedback, the control policy struggles to maintain a stable flight height due to the unexpected heavy payload. In contrast, our method, which integrates predictive error feedback, achieves robust tracking despite the downward force from the additional payload.

\section{Conclusion}
\label{sec:conc}

In this paper, we propose Neural Internal Model Control (NeuralIMC), a framework that bridges model-based control with RL-based neural control. By leveraging body dynamics and a white-box rigid-body transformation, we avoid high-dimensional nonlinearity, using a simplified predictive model in a closed-loop feedback structure. Model-free RL enables the policy to react to predictive errors. NeuralIMC outperforms state-of-the-art controllers on quadrotors and quadrupedal robots. Real-world tests on a heavily loaded quadrotor demonstrate robustness to disturbances and dynamic variations, highlighting its potential for interpretable and adaptive learning-based control. For the future work, we'd like to conduct more expanded real-world experiments includes testing on diverse robots and improving disturbance training~\cite{long2024learning}.

%
\IEEEpeerreviewmaketitle

\ifCLASSOPTIONcaptionsoff
  \newpage
\fi



%
\bibliographystyle{IEEEtran}
\bibliography{ref}

\end{document}